%% file: roam.tex
\documentclass[10pt,twocolumn,letterpaper]{article}

\usepackage{cvpr}
\usepackage{times}
\usepackage{epsfig}
\usepackage{graphicx}
\usepackage{amsmath}
\usepackage{amssymb}

\usepackage{comment}
\usepackage{bm}
\usepackage{multirow}
\usepackage{subfigure}
\usepackage{microtype}
\usepackage{color}
\usepackage{listings}
\usepackage{algorithm}
\usepackage{algpseudocode}
\usepackage{mathrsfs}
\usepackage{caption}
\usepackage{stmaryrd}
\usepackage{xcolor}
\usepackage[compact]{kmisc}

\input{text/macros}

\newcommand{\acro}{{\roam}}

\usepackage[pagebackref=true,breaklinks=true,letterpaper=true,colorlinks,bookmarks=false]{hyperref}

\cvprfinalcopy % *** Uncomment this line for the final submission

\def\cvprPaperID{1953} % *** Enter the CVPR Paper ID here

\begin{document}

%%%%%%%%% TITLE
\title{ROAM: a Rich Object Appearance Model with Application to Rotoscoping}

\author{
\begin{tabular*}{0.85\linewidth}{@{\extracolsep{\fill}}cccc}
	Ondrej Miksik$^{1}$\thanks{Assert joint first authorship.} & Juan-Manuel P\'erez-R\'ua$^{2*}$ & Philip H. S. Torr$^1$ & Patrick P\'erez$^2$ \\
	\multicolumn{4}{c}{$^1$University of Oxford}\\
	\multicolumn{4}{c}{$^2$Technicolor Research \& Innovation}
\end{tabular*}
}

\maketitle
%\thispagestyle{empty}

%%%%%%%%% ABSTRACT
\begin{abstract}
Rotoscoping, the detailed delineation of scene elements through a video shot, is a painstaking task of tremendous importance in professional post-production pipelines. While pixel-wise segmentation techniques can help for this task, professional rotoscoping tools rely on parametric curves that offer the artists a much better interactive control on the definition, editing and manipulation of the segments of interest. Sticking to this prevalent rotoscoping paradigm, we propose a novel framework to capture and track the visual aspect of an arbitrary object in a scene, given a first closed outline of this object. This model combines a collection of local foreground/background appearance models spread along the outline, a global appearance model of the enclosed object and a set of distinctive foreground landmarks. The structure of this rich appearance model allows simple initialization, efficient iterative optimization with exact minimization at each step, and on-line adaptation in videos. We demonstrate qualitatively and quantitatively the merit of this framework through comparisons with tools based on either dynamic segmentation with a closed curve or pixel-wise binary labelling. 

\end{abstract}

%\begin{figure*}
%	\vspace{-0.5cm}
%	\includegraphics[trim=0 13.5cm 0 0,clip,width=1\linewidth]{fig/teaser3.pdf}\\
%	\caption{\textbf{ROAM for video object cutout}. Starting from an initial outline provided by the user (left), the proposed appearance model allows the \emph{automatic} segmentation of a complex object through a video sequence. 
%	}
%	\label{fig:teaser}
%\end{figure*}

%%%%%%%%% BODY TEXT
\input{text/intro.tex}

\input{text/related.tex}

\input{text/model.tex}
\input{text/videocut.tex}

\input{text/results.tex}
\input{text/conclusion_new.tex}

{\small
\bibliographystyle{ieee}
\bibliography{roam}
}

\end{document}

%% file: text/macros.tex
\newcommand{\roam}{{\sc Roam}}
\newcommand{\snap}{{\sc Video SnapCut}}
\newcommand{\roto}{{\sc Rotobrush}}
\newcommand{\rotop}{{\sc Roto++}}
\newcommand{\after}{{\sc Ae}}
\newcommand{\cpc}{{\sc Cpc}}
\newcommand{\davis}{{\sc Davis}}
\newcommand{\jump}{{\sc JumpCut}}
\newcommand{\grab}{{\sc GrabCut}}
\newcommand{\glob}{\mathrm{glob}}
\newcommand{\loc}{\mathrm{loc}}

\newcommand\ignore[1]{}

\def \path{\bp C}

\newcommand{\bfe}{{\mathbf{e}}}

\newcommand{\bfp}{{\mathbf{p}}}
\newcommand{\bfq}{{\mathbf{q}}}

\newcommand{\bfx}{{\mathbf{x}}}
\newcommand{\bfy}{{\mathbf{y}}}

\newcommand{\bfI}{{\mathbf{I}}}

\newcommand{\mubf}{\boldsymbol{\mu}}

\newcommand{\calI}{{\mathcal{I}}}

\newcommand{\calO}{{\mathcal{O}}}
\newcommand{\calP}{{\mathcal{P}}}

\newcommand{\calX}{{\mathcal{X}}}
\newcommand{\calY}{{\mathcal{Y}}}

%% file: text/intro.tex
\section{Introduction}
\label{sec:introduction}

Modern high-end visual effects (vfx) and post-production rely on complex workflows whereby each shot undergoes a succession of artistic operations. Among those, rotoscoping is probably the most ubiquitous and demanding one \cite{bratt2011rotoscoping,li2016roto++}. Rotoscoping amounts to outlining accurately one or several scene elements in each frame of a shot. This is first a key operation for compositing \cite{wright2006digital} (insertion of a different background, whether natural or synthetic), where it serves as an input to subsequent operations such as matting and motion blur removal.\footnote{The use of blue or green screens on set can ease compositing but remains a contrived set-up. Even if accessible, such screens lead to chroma-keying and de-spilling operations that are not trivial and are not suited to all foreground elements, thus rotoscoping remains crucial.} Rotoscoping is also a pre-requisite for other important operations, such as object 
color grading, rig removal and new view synthesis, with large amounts of elements to be handled in the latter case.

Creating such binary masks is a painstaking task accomplished by trained artists. It can take up to several days of work for a complex shot of only a few seconds, using dedicated tools within video editing softwares like \href{https://www.silhouettefx.com/}{Silhouettefx}, Adobe \href{http://www.adobe.com/fr/products/aftereffects.html}{After Effect}, Autodesk \href{http://www.autodesk.com/products/flame-family/overview}{Flame} and The Foundry \href{https://www.thefoundry.co.uk/products/nuke/}{Nuke}. As discussed in \cite{li2016roto++}, professional roto artists use mostly tools based on \textit{roto-curves}, \ie, parametric closed curves that can be easily defined, moved and edited throughout shots. By contrast, these artists hardly use brush-based tools, even if empowered by local appearance modelling, graph-based regularization and optic flow-based tracking as After Effect's \roto. 

Due to its massive prevalence in professional workflows, we address here rostoscoping in its closed contour form, which we aim to facilitate. Roto curves being interactively placed in selected keyframes, automation can be sought either at the key-frame level (reducing the number of user's inputs) or at the tracking level (reducing the number of required key-frames). In their recent work, Li \etal \cite{li2016roto++} offers with \rotop, a tool that helps on both fronts, thanks to an elegant shape modelling. In present work, we explore a complementary route that focuses on the automatic tracking from a given keyframe. In essence, we propose to equip the roto-curve with a rich, adaptive modelling of the appearance of the enclosed object. This model, coined {\sc roam} for Rich Online Appearance Model, combines in a flexible ways various appearance modelling ingredients: (i) Local foreground/background color modelling, in the spirit of \snap~\cite{bai2009video} but attached here to the roto-curve; (ii) Fragment-based modelling to handle large displacements and deformations and (iii) Global appearance modelling, which has proved very powerful in binary segmentation with graph cuts, \eg in \cite{boykov2001interactive}. 

As demonstrated on recent benchmarks, \roam~outperforms state-of-art approaches when a single initial roto-curve is provided. It is in particular less prone than After Effect \roto~ to spurious changes of topology that lead to eventual losses, and more robust than \rotop~\cite{li2016roto++} in the absence of additional user inputs. This robustness makes it appealing to facilitate rotoscoping, either as a standalone tool, or combined with existing curve-based tool such as \rotop.

%% file: text/related.tex
%\section{On modeling visual appearance}
\section{Related work and motivation}
\label{sec:related}

Rotoscoping is a form of interactive ``video object''\footnote{Throughout, ``video object'', or simply ``object'', is a generic term to designate a scene element of interest and the associated image region in the video} segmentation. As such, the relevant literature is vast. For sake of brevity, we focus mostly our discussion on works that explicitly target rotoscoping or very similar scenarios.
%, before briefly reviewing some key tools for modeling shape and appearance of arbitrary video objects.  

\subsection{Rotoscoping and curve-based approaches}

Li \etal \cite{li2016roto++} recently released a very detailed study of professional rotoscoping workflow. 
They first establish that trained artists mostly use parametric curves such as Bezier splines to delineate objects of interest in key-frames, ``track'' them from one frame to the next, edit them at any stage of the pipeline and, last but not least, to pass them in a compact and manipulable format to the next stage of the fvx pipeline, \eg, to the compo-artists.
Professional rotoscoping tools such as Silhouettefx, Blender, Nuke or Flame are thus based on parametric curves, which can be either interpolated between key-frames or tracked with a homographic ``planar tracker'' when suitable. 
Sticking to this ubiquitous workflow, the authors propose \rotop~ to speed it up. 
Bezier roto-curves defined by the artist in the selected key-frames allow the real-time learning of a non-linear low-dimensional shape space based on a Gaussian process latent variable model. 
Shape tracking between key-frames, as well as subsequent edits, are then constrained within this smooth manifold (up to planar transforms), with substantial gains in work time. 
Our work is fully complementary to \rotop: while \roam~does not use a strong shape prior in its current form, it allows to capture the dynamic appearance of the video object, something that \rotop~ does not address. 
Note that the annotation (ground-truth roto-curves) of the rotoscoping dataset that constitutes another contribution of \cite{li2016roto++} is not yet available, as per submission time. 

In their seminal rotoscoping work, Agarwala \etal \cite{agarwala2004keyframe} proposed a complete interactive system to track and edit Bezier roto-curves. 
It relies on the popular active contour framework \cite{blake2000active,kass1988snakes}: a curve, parametrized by control points, finely discretized and equipped with a second-order smoothness prior is encouraged to evolve smoothly and snap to strong edges of the images. 
Their energy-based approach also uses local optical flow along each side of the shape's border. 
In contrast to this work, our approach offers a richer local appearance modelling along the roto-shape as well as additional intra-object appearance modelling.

Similarly to \cite{agarwala2004keyframe}, Lu \etal \cite{lu2016coherent} recently introduced an interactive object segmentation system called ``coherence parametric contours'' (\cpc), which combines planar tracking with active contours. 
Our system includes similar ingredients, with the difference that the planar tracker is subsumed by a fragment-based tracker and that the appearance of the object and of its close surrounding is also captured and modeled. 
We demonstrate the benefits of these additional features on the evaluation dataset introduced by Lu \etal~\cite{lu2016coherent}.

\subsection{Masks and region-based approaches}

Other notable works on interactive video segmentation address directly the problem of extracting binary masks, \ie labelling pixels of non-keyframes as foreground or background. 
As discussed in \cite{li2016roto++,lu2016coherent}, a region-based approach is less compatible with professional rotoscoping, yet provides powerful tools. 
Bai \etal \cite{bai2009video} introduced \snap, which lies at the heart of After Effect's \roto. 
Interaction in \snap~ is based on foreground/background brushes, following the popular scribble paradigm of Boykov and Jolly \cite{boykov2001interactive}. 
The mask available in a given frame is tracked to the next frame through the propagation of local windows that straddle its border. 
Each window is equipped with a local foreground/background color model and a local shape template, both updated through time. After propagation along an object-centric optical-flow, these windows provide suitable pixel-wise unaries that are fed to a classic graph-cut. 
This approach provides a powerful way to capture on-the-fly local color models and combine them adaptively with some shape persistence. 
However, being based on graph-cut (pixel-wise labelling), \roto~ can be penalized by its extreme topology flexibility: as will be showed in the experiments, rapid movements of the object, for instance, can cause large spurious deformations of the mask that can eventually lead to complete losses in the absence of user intervention.  
In \roam, we take inspiration from the local color modelling at the object's border and revisit it in a curve-based segmentation framework that allows tighter shape control and easier subsequent interaction.

More recently, Fan \etal introduced \jump~\cite{fan2015jumpcut}, another mask-based approach where frame-to-frame propagation is replaced by mask transfer from the key-frame(s) to distant frames. 
This long-range transfer leverages dense patch correspondences computed over the inside and outside of the known mask, respectively. 
The transfered mask is subsequently refined using a standard level set segmentation (region encoded via a spatial map). 
A salient edge classifier is trained online to locate likely fragments of object's new silhouette and drive the level set accordingly. 
They reported impressive results with complex deformable objects going through rapid changes in scene foreground. 
However, similarly to \roto, this agility might also become a drawback in real rotoscoping scenarios, as is the lack of shape parametrization. 
Also, the underlying figure/ground assumption (the object is moving distinctly in front of a background) is not met in many cases, \eg rotoscoping of a static scene element or of an object in a dynamic surrounding.

%% file: text/model.tex
% ---------------------------------------------------------------
\section{Introducing ROAM}
\label{sec:model}

Our model consists of a graphical model with the following components: (i) A closed curve that defines an object and a collection of local foreground/background\footnote{``Foreground/background'' terminology, ``fg/bg'' in short, merely refers here to inside and outside of the roto-curve; it does not imply that the object stands at the forefront of the 3D scene with a background behind it.} appearance models along it; (ii) A global appearance model of the enclosed object; (iii) A set of distinctive object's landmarks.
While the global appearance model captures image statistics as in graph-cut approaches \cite{boykov2001interactive, rother2004grabcut}, it is the set of local fg/bg appearance models placed along the boundary that enables accurate object delineation. 
The distinctive object's landmarks organized in a star-shaped model (Fig.~\ref{fig:roam}, left) help to prevent the contour from sliding along itself and to control the level of non-rigid deformations.
The landmarks are also used to robustly estimate a rigid transformation between the frames to ``pre-warp'' the contour, which significantly speeds-up the inference.
In addition, the control points of the roto-curve, as well as the local fg/bg models and the landmarks are maintained through time, which provides us with different types of temporal correspondences.

\begin{figure}[t]
\begin{center}
\fbox{\includegraphics[width=0.80\linewidth]{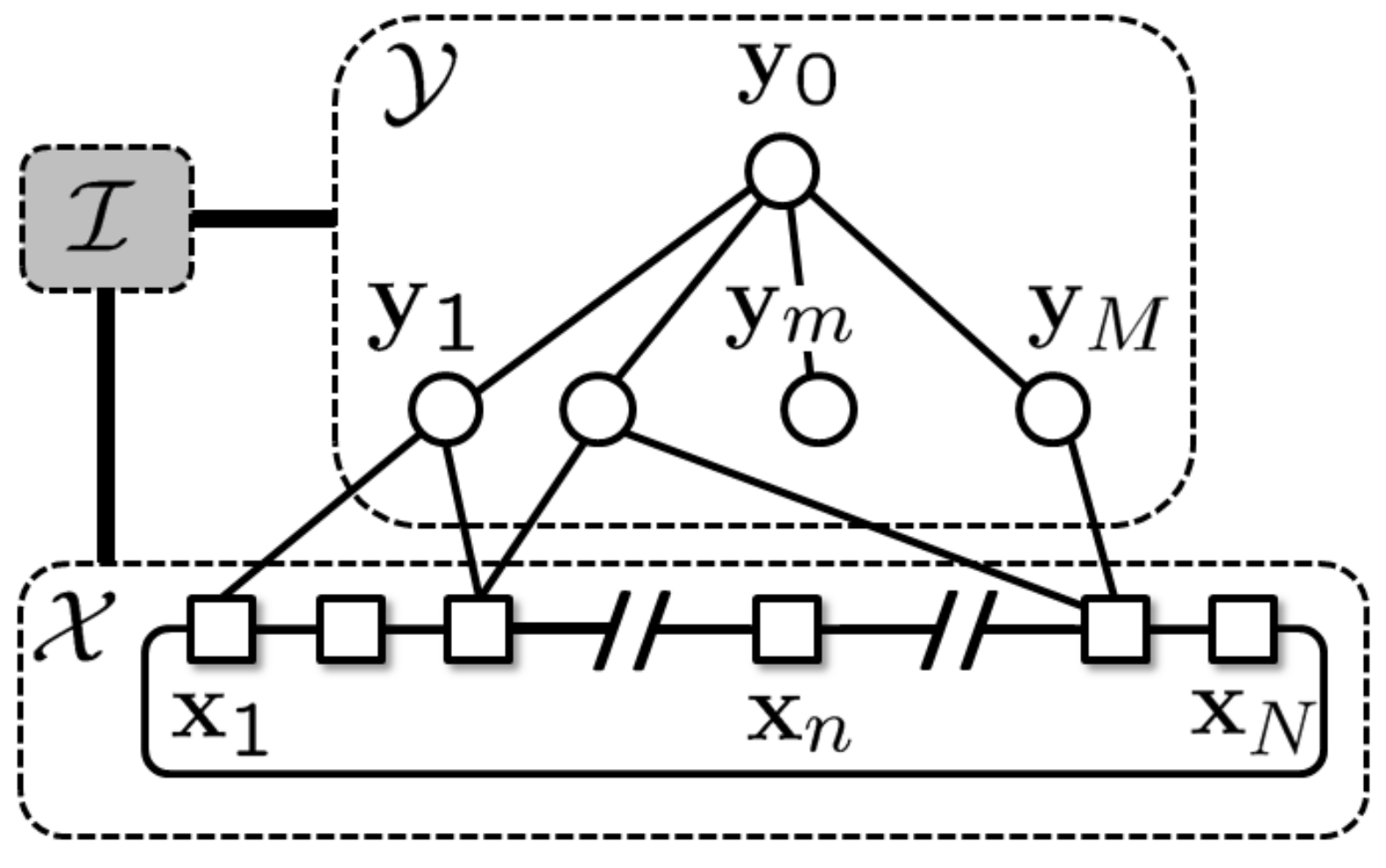}}
\end{center}
\caption{{\bf Graphical model of \acro~}. In joint model defined by energy $E(\calX,\calY;\calI)$ in~\eqref{eq:E}, contour node variables (white squares) form a closed 1-st order chain conditioned on image data (grey box) and landmark variables (white circles), the latter variables forming a shallow tree conditioned on all others.} 
\label{fig:graph}
\end{figure}

% ---------------------------------------------------------------
%\subsection{Model}

Given color image $\calI = \{\bfI_{\bfp}\}_{\bfp\in\calP}$, a conditional graphical model (Fig.~\ref{fig:graph}) is defined through the energy function
\begin{equation}
E(\calX, \calY; \calI) := E^C(\calX; \calI) + E^L(\calY; \calI) + E^J(\calX, \calY),
\label{eq:E}
\end{equation}
where $E^C$ and $E^L$ depend only on the roto-curve configuration $\calX$ and the landmarks configuration $\calY$ respectively, and $E^J$ links the two together (independently of the image).
In the following, we describe these three energy terms in detail.

% ---------------------------------------------------------------
\subsection{Curve-based modelling: $E^C$}

\begin{figure}[t]
\centering
\includegraphics[width=1\linewidth]{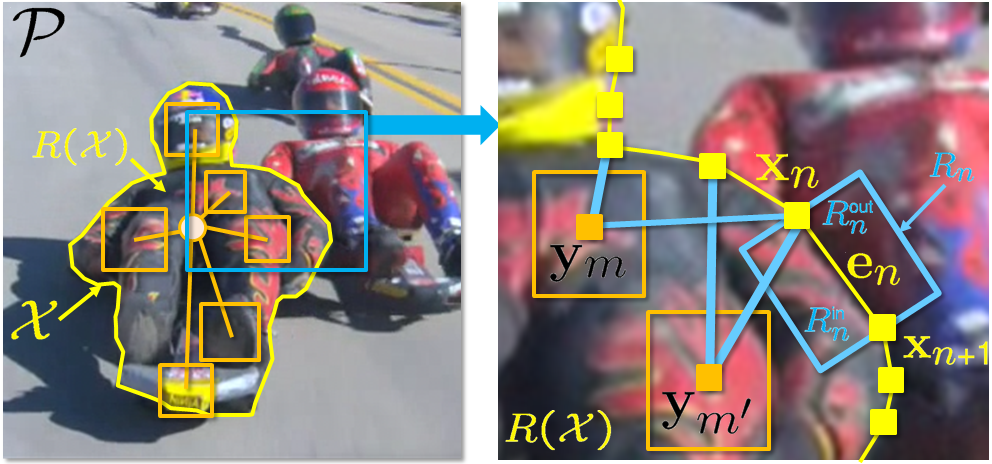}
\caption{{\bf Structure and notations of proposed model}. (Left) A simple closed curve $\calX$ outlines the object region $R(\calX)$ in the image plane $\calP$. Several landmarks, forming a star-shaped graphical model, are defined in this region. (Right) Each edge $\bfe_n$ of the closed polyline defines a region $R_n$ that staddles $R(\calX)$; each node $\bfx_n$ of the polyline is possibly connected to one or several landmarks.}
\label{fig:roam}
\end{figure}

%We use a parametric representation of the closed roto-shape. 
While Bezier splines are a popular representation for rotoscoping \cite{agarwala2004keyframe,li2016roto++}, we simply consider polygonal shapes here: 
roto-curve $\calX$ is a polyline with $N$ vertices $\bfx_1\ldots \bfx_N\in\mathbb{Z}^2$ and $N$ non-intersecting edges $\bfe_{n}=(\bfx_{n},\bfx_{n+1})$, where $\bfx_{N+1}$ stands for $\bfx_1$, \ie the curve is closed.
Given an orientation convention (\eg clockwise), the interior of this curve defines a connected subset $R(\calX)\subset\calP$ of the image pixel grid (Fig.~\ref{fig:roam}, left), which will be denoted $R$ in short when allowed by the context.

Energy $E^C$ is composed of two types of edge potentials $\psi_n^{\loc}$ and $\psi_n^{\glob}$ that relate to local and global appearance respectively:
\begin{equation}
\begin{split}
E^C(\calX; \calI) &:= \sum_{n=1}^N  [\psi_n^{\loc}(\bfe_n) + \psi_n^{\glob}(\bfe_n)].
\label{eq:EC}
\end{split}
\end{equation}
As with classic active contours \cite{kass1988snakes}, the first type of potential will encapsulate both a simple $\ell_2$-regularizer that penalizes stretching and acts as a curve prior (we are not using second-order smoothing in the current model), and a data term that encourages the shape to snap to strong edges. It will in addition capture color likelihood of pixels on each side of each edge via local appearance models.   
The second set of potentials results from the transformation of object-wise color statistics (discrete surface integral) into edge-based costs (discrete line integrals).

Note that, since we do not impose any constraint on the various potentials, the one specified below could be replaced by more sophisticated ones, \eg using semantic edges~\cite{sketchtokens} instead of intensity gradients, or using statistics of convolutional features~\cite{richfeaturehierarchies} rather than color for local and global appearance modelling. 

% ---------------------------------------------------------------
{\bf Local appearance model.}
Each edge $\bfe_n$ is equipped with a local appearance model $p_n = (p_n^{\mathrm f}, p_n^{\mathrm b})$ composed of a fg/bg colour distribution and of a rectangular support $R_n$, with the edge as medial axis and a fixed width in the perpendicular direction (Fig.~\ref{fig:roam}, right). 
Denoting $R_n^{\mathrm{in}}$ and $R_n^{\mathrm{out}}$ the two equal-sized parts of $R_n$ that are respectively inside and outside $R$, we construct a simple edge-based energy term (the smaller, the better) that rewards edge-configurations such that: colours in $R_n^{\mathrm{in}}$ (resp. $R_n^{\mathrm{out}}$) are well explained by model $p_n^{\mathrm f}$ (resp. $p_n^{\mathrm g}$) and edge $\bfe_n$ is short and goes through high intensity gradients: 
\begin{align}
\psi_n^{\loc}(\bfe_n)\ := &
- \sum_{\bfp\in R_n^{\mathrm{in}}} \ln p_n^{\mathrm f}(\bfI_{\bfp}) \label{eq:e_contour}
- \sum_{\bfp\in R_n^{\mathrm{out}}} \ln p_n^{\mathrm b}(\bfI_{\bfp})\\
& + \sum_{\bfp\in\bfe_n}  \big( \mu \|\bfx_{n+1} - \bfx_n\|^2 - \lambda\|\nabla \calI(\bfp))\|^2 \big),\nonumber
\end{align}
with $\mu$ and $\lambda$ two positive parameters.
%\ONDRA{I think the gradient term should be turned into ``edge uncertainty''}

% ---------------------------------------------------------------
{\bf Global appearance model.}
A global appearance model captures image statistics over the object's interior.
As such, it also helps pushing the roto-curve closer to the object's boundary, especially when local boundary terms are not able to explain foreground and background reliably.
Defining $p_0=(p_0^{\mathrm f},p_0^{\mathrm b})$ the global fg/bg colour distribution, the bag-of-pixel assumption allows us to define region energy term
\begin{equation}
\sum_{\bfp\in R} \ln \frac{p_0^{\mathrm b}(\bfI_{\bfp})}{p_0^{\mathrm f}(\bfI_{\bfp})}.
\end{equation}
This discrete region integral can be turned into a discrete contour integral using one form of discrete Green theorem~\cite{tang1982discrete}.
Using horizontal line integrals for instance, we get% (Fig.~\ref{fig:green}): 
\begin{equation}
\sum_{\bfp\in R} \ln \frac{p_0^{\mathrm b}(\bfI_{\bfp})}{p_0^{\mathrm f}(\bfI_{\bfp})} = \sum_{n=1}^N \underbrace{\sum_{\bfp\in\bfe_n} \alpha_n(\bfp) Q(\bfp)}_{:=\psi_n^{\glob}(\bfe_n)},% = \sum_{n=1}^N \psi_{\glob}(\bfe_n),
\label{eq:green}
\end{equation}
where $Q(\bfp)=\sum_{``\bfq\leqslant\bfp''}\ln (p_0^{\mathrm b}(\bfI_{\bfp})/p_0^{\mathrm f}(\bfI_{\bfp}))$ is the discrete line integral over pixels to the left of $\bfp$ on the same row, and $\alpha_n(\bfp)\in\{-1,+1\}$ depends on the direction and orientation, relative to curve's interior, of the oriented edge $\bfe_n$. 
In \eqref{eq:green}, the second sum in r.h.s. is taken over the pixel chain resulting from the discretization of the line segment $[\bfx_{n},\bfx_{n+1}]$ with the final vertex excluded to avoid double-counting. 

% ---------------------------------------------------------------
\subsection{Landmark-based modelling: $E^L$}

Our model also makes use of a set $\calY$ of $M$ distinctive landmarks $\bfy_1 \ldots \bfy_M \in R(\mathcal{X})$ detected inside the object of interest. Similarly to pictorial structures \cite{felzenszwalb2010object}, these landmarks form the leaves of a star-shaped graphical model\footnote{The star shape is used for its simplicity but could be replaced by another tree-shaped structure.} with a virtual root-node $\bfy_0$.
This part of the model is defined by % virtual root-node potential $\phi_0(\bfy_0)$ (taken uniform), 
leaf potentials $\phi_m(\bfy_m)$ and leaf-root potentials $\varphi_m(\bfy_0, \bfy_m)$:
\begin{equation}
E^L(\calY; \calI):= \sum_{m=1}^M \phi_m(\bfy_m) + \sum_{m=1}^M \varphi_m (\bfy_0, \bfy_m).
\label{eq:EL}
\end{equation}
Each landmark is associated with a model, \eg a template or a filter, that allows the computation of a matching cost at any location in the image. The leave potential $\phi_m(\bfy_m)$ corresponds to the negative matching cost for $m$-th landmark.
The pairwise potentials $\varphi_{m}$ penalize the difference in $\ell_2$-norm between the current configuration and the one, $\hat{\calY}$, estimated in previous frame:
\begin{equation}
\varphi_{m}(\bfy_0, \bfy_m) = \frac{1}{2} \|\bfy_m-\bfy_0-\hat{\bfy}_m+\hat{\bfy}_0\|^2.
\end{equation}
%

% ---------------------------------------------------------------
\subsection{Curve-landmarks interaction: $E^J$}

The joint energy $E^J(\calX, \calY)$ captures correlation between object's outline and object's landmarks.
Based on proximity, shape vertices and landmarks can be associated. Let denote $n\sim m$ the pairing of vertex $\bfx_n$ with landmark $\bfy_m$. Energy term $E^J$ decomposes over all such pairs as:
\begin{equation}
E^J(\calX,\calY) = \sum_{n\sim m} \xi_{nm}(\bfx_n,\bfy_m).
\label{eq:EJ}
\end{equation}
For each pair $n\sim m$, 
%the vertex-to-landmark shift vector being $\mubf_{mn}$ 
%and a covariance matrix $\Sigma_{mn}$ as part of the model, an 
the interaction potential is defined as:
\begin{equation}
\xi_{mn}(\bfx_n,\bfy_m) = \frac{1}{2}\|\bfx_n-\bfy_m-\mubf_{mn}\|^2,
\label{eq:xi_mn}
\end{equation}
where $\mubf_{mn}$ is the landmark-to-vertex shift vector in the first image. 

%% file: text/videocut.tex
\section{Using ROAM}
\label{sec:videocut}

{\bf Sequential alternate inference.} 
Using \acro~ to outline the object of interest in a new image amounts to solving the discrete optimization problem:
\begin{equation}
\min_{\calX,\calY} E(\calX,\calY;\calI),
\end{equation} 
where $E$ is defined by (\ref{eq:E}) and depends on previous \mbox{curve/landmarks} configuration $(\hat{\calX},\hat{\calY})$ through several of its components.  
Despite this problem could be formulated as an integer linear program, we opt for simpler alternating optimization with exact minimization at each step which converges within a few iterations.% (Sec. \ref{sec:results}).

In the first step, we fix the roto-curve $\calX$ and find the best configuration of landmarks $\calY$ using dynamic programming. 
Exact solution for such a problem can be obtained in two passes, solving exactly
\begin{equation}
\begin{split}
\min_{\bfy_0} \min_{\bfy_{1:M}}  \sum_{m=1}^M  \bigg( & \phi_m(\bfy_m) + \varphi_m(\bfy_0,\bfy_m) \\
& + \sum_{n\sim m} \xi_{mn}(\bfx_n,\bfy_m) \bigg).
\end{split}
\end{equation}
Default implementation leads to complexity $\calO(MS^2)$, with $S$ the size of individual landmark state-spaces, \ie the number of possible pixel positions allowed for each. However, the quadratic form of the pairwise terms allows making it linear in the number of pixels, \ie $\calO(MS)$, by resorting to generalized distance transform \cite{felzenszwalb2010object}.  

In the second step, we fix the landmarks $\calY$ and find the best configuration of contour $\calX$. This is a classic first-order active contour problem. Allowing continuous values for nodes coordinates, a gradient descent can be conducted, with all nodes being moved simultaneously at each iteration. We prefer the discrete approach, whereby only integral positions are allowed and dynamic programming can be used \cite{amini1990using}. In that formulation, exact global inference is theoretically possible, but with a prohibitive complexity of $\calO(NP^3)$, where $P=\mathrm{card}(\calP)$ is the number of pixels in images. We follow the classic iterative approach that considers only $D$ possible moves $\Delta\bfx$ for each node around its current position. For each of the $D$ positions of first node $\bfx_1$, Viterbi algorithm provides the best moves of all others in two passes and with complexity $\calO(ND^2)$. Final complexity is thus $\calO(ND^3)$ for each iteration of optimal update of previous contour, solving:
\begin{align}
\min_{\Delta\bfx_1}\min_{\Delta\bfx_{2:N}} \sum_{n=1}^N  
\bigg(  \nonumber &
\psi_n^{\loc}(\bfe_n+\Delta\bfe_n) + \psi_n^{\glob}(\bfe_n+\Delta\bfe_n) \\
& + \sum_{m\sim n} \xi_{mn}(\bfx_n+\Delta\bfe_n,\bfy_m)
\bigg).
\end{align}
Note that sacrifying optimality of each update, the complexity could even been reduced as much as $\calO(ND)$ \cite{williams1992fast}.

Given some initialization for $(\calX,\calY)$, we thus alternate between two \textit{exact} block-wise inference procedures. This guaranties convergence toward a local minima of joint energy $E(\calX,\calY;\calI)$. Also, the complexity of each iteration is linear in the number of vertices and landmarks, linear in the number of pixels, and cubic in the small number of allowed moves for a curve's vertex.

{\bf Online learning of appearance models.} Local fg/bg color models $p_n$s and global color model $p_0$ are GMMs. Given the roto-curve in the initial frame, these GMMs are first learned over region pairs $(R_n^{\mathrm{in}},R_n^{\mathrm{out}})$s and $(R,\calP \setminus R)$ respectively and subsequently adapted through time using Stauffer and Grimson's classic technique \cite{stauffer1999adaptive}.

{\bf Selection and adaption of landmarks.} A pool of distinctive landmarks is maintained at each instant. They can be any type of classic interest points. In order to handle texture-less objects, we use maximally stable extremal regions (MSERs) \cite{matas2004robust}. Each landmark is associated with a correlation filter whose response over a given image area can be computed very efficiently \cite{henriques2015kernelized}. At any time, landmarks whose filter response is too ambiguous are deemed insufficiently discriminative and removed from the current pool in the same way tracker loss is monitored in \cite{henriques2015kernelized}. The collection is re-populated through new detections. Note that correlation filters can be computed over arbitrary features and kernelized \cite{henriques2015kernelized}; for simplicity, we use just grayscale features without kernel function. 

{\bf Allowing topology changes.} Using a closed curve is crucial to comply with rotoscoping workflows and allows the definition of a rich appearance model. Also, it prevents abrupt changes of topology. While this behavior is overall beneficial (See \S \ref{sec:results}), segmenting a complete articulated 3D object as in Fig. 1 might turn difficult. Roto-artists naturally handle this by using multiple independent roto-curves, one per meaningful part of the object. As an alternative for less professional, more automatic scenarios, we propose to make \roam~ benefit from the best of both worlds: standard graph-cut based segmentation \cite{boykov2001interactive}, with its superior agility, is used to \textit{propose} drastic changes to current curve, if relevant. Space-dependent unaries are derived in ad-hoc way from both global and local color models and combined with classic contrast-dependent spatial regularization.\footnote{Note that this instrumental energy is too poor to compete on its own with proposed model, but is a good enough proxy for the purpose of proposing possibly interesting new shapes at certain instants. It is also very different from the one in the final graph-cut of \snap~where unaries are based on the already computed soft segmentation to turn it into a hard segmentation. Also, graph-cut segmentation is the final output in \snap, unless further interaction is used, while we only use it to explore alternative topologies under the control of our joint energy model.} The exact minimizer of this instrumental cost function is obtained through graph-cut (or its dynamic variant for efficiency \cite{kohli2007dynamic}) and compared to the binary segmentation associated to the current shape $\calX$. At places where the two differ significantly, a modification of current configuration (displacement, removal or addition of vertices) is proposed and accepted if it reduces the energy $E(\calX,\calY;\calI)$.

%% file: text/results.tex
\section{Results}
\label{sec:results}

We now report experimental comparisons that focus on the minimum input scenario: an initial object selection (curve or mask, depending on the tool) is provided to the system and automatic object segmentation is produced in the rest of the shot.\footnote{Video results are available at \url{https://youtu.be/UvO7IacS9pQ}} We do not consider additional user interactions.

% ---------------------------------------------------------------
{\bf Datasets.}
We evaluate our approach on the recent \cpc~rotoscoping dataset \cite{lu2016coherent}. It contains 9 videos consisting of 60 to 128 frames which represent typical length of shots for rotoscoping.
These sequences were annotated by professional artists using standard post-production tools.
We also provide qualitative results on shots from the \rotop~ \cite{li2016roto++} dataset (Fig. \ref{fig:results_rotop}) for which the authors have not released the ground-truth yet, as well as from the \snap~dataset \cite{bai2009video} (Fig. \ref{fig:results_misc}). 

In addition to that, we use the \davis~video segmentation dataset \cite{perazzi2016a} which comprises 
50 challenging sequences with a wide range of difficulties: large occlusions, long-range displacements, non-rigid deformations, camouflaging effects and complex multi-part objects. 
Let us note that this dataset is intended to benchmark pixel-based video segmentation methods, not rostocoping tools based on roto-curves.

% ---------------------------------------------------------------
{\bf Evaluation Metrics.}
We use standard video segmentation evaluation metrics and report the average \textit{accuracy}, \ie, the proportion of ground-truth pixels that are correctly identified, and the more demanding average \textit{intersection-over-union} (IoU), \ie, the area of the intersection of ground-truth and extracted objects over the area of their union. 
We also report runtimes and evolution of IoU as sequences proceed. %as the number of processed frames increases. 

\begin{table*}[tb]
\caption{Quantitative comparisons on \davis~dataset}
\small
\setlength{\tabcolsep}{2pt}
\centering
\begin{tabular}{c|c|c|c|c|c|c|c|c|c}
\hline
 & \multicolumn{3}{c|}{Average Accuracy} & \multicolumn{3}{c|}{Average IoU} & \multicolumn{3}{c}{Time / frame (s)}\\
Method & Validation set & Training set &~Full set~& Validation set & Training set & ~Full set~ & min & max & avg \\
\hline
\hline
\grab~\cite{rother2004grabcut} + Tracker \cite{henriques2015kernelized} & 0.896 & 0.914 & 0.907 & 0.277 & 0.296 & 0.289 & 0.405 &  0.675 & 0.461  \\
\jump~\cite{fan2015jumpcut} & \bf{0.952} & \bf{0.957} & \bf{0.956} & \underline{0.570} & \bf{0.632} & \bf{0.616} & - & - & -\\
\after~\roto~\cite{bai2009video} & \underline{0.951} & 0.942 & 0.946 & 0.533 & 0.479 & 0.500 & - & - & - \\
\rotop~(single keyframe)\cite{li2016roto++} & 0.910 & 0.922 & 0.917 & 0.248 & 0.310 & 0.284 & - & - & 0.118 \\
\rotop~(two keyframes) \cite{li2016roto++} & 0.925 & 0.933 & 0.930 & 0.335  & 0.394 & 0.358 & -	& - & 0.312\\
\hline
\acro: Baseline Conf.  & 0.930 & 0.937 & 0.932 & 0.358 & 0.385 & 0.377 & 0.028 & 0.122 & 0.056 \\ %5-4 (GRAD+NORM+9)
\acro: Lean Conf.  & 0.935 & 0.937 & 0.936 & 0.409 & 0.417 & 0.412 & 0.258 & 0.622 & 0.356 \\ %6-5 (SC+GRAD+NORM)
\acro: Medium Conf. & 0.942 &  0.952 & 0.948 &  0.532  & 0.591 & 0.564 & 0.487 & 2.251 & 0.971 \\%SC+GRAD+NORM+LAND
\acro: Full Conf.  & \bf{0.952} & \underline{0.956} & \underline{0.953} & \bf{0.583} & \underline{0.624} & \underline{0.615} & 1.981 & 9.450 & 5.639 \\%SC+SEMANTIC GRAD+NORM+LAND+REPAR
\hline
\end{tabular}
\label{tab:comp_iou_davis}
\vspace{0.15cm}
\end{table*}

% ---------------------------------------------------------------
{\bf Baselines.} 
We compare with several state-of-the-art methods. 
Our main comparison is with recent approaches that rely on a closed-curve, \ie, \cpc~\cite{lu2016coherent} and \rotop~\cite{li2016roto++}. 
We initialize all methods with the same object %in the first frame 
and measure their performance over the rest of each sequence.
Since \rotop~ requires at least two key-frames to benefit from its online shape model, 
we report scores with letting the method access the ground-truth of the last frame as well. 
%report scores using two keyframes for this method (first and last frames of the sequence).
%
We also run it with the initial keyframe only, a configuration in which \rotop~ boils down to Blender planar tracker. 

In addition to that, we also compare with two approaches based on pixel-wise labelling:
\jump~\cite{fan2015jumpcut} and \snap~\cite{bai2009video} as implemented in After Effect \roto.
As a naive baseline, we use a combination of a bounding-box tracker~\cite{henriques2015kernelized} and \grab~\cite{rother2004grabcut}.

\begin{table}[tb]
\caption{Quantitative evaluation on \cpc~dataset ($^*$: partial evaluation only, see text)} %For each method, the average intersection-over-union, IoU (the higher, the better) is reported.
\setlength{\tabcolsep}{2pt}
\small
\begin{tabular}{c|c|c|c|c|c}
\hline
 & \multicolumn{1}{c|}{Avg. } & \multicolumn{1}{c|}{Avg.} & \multicolumn{3}{c}{Time (s) / frame}\\
Method & Accuracy &  IoU & min & max & avg \\
\hline
\hline
\grab~\cite{rother2004grabcut} + Tracker \cite{henriques2015kernelized} & .891 & .572 &  0.394  & 0.875 & 0.455\\
\after~\roto~\cite{bai2009video} & .992  & .895 & --- & --- & ---\\
\rotop (1 keyframe)~\cite{li2016roto++} & .969 & .642 & --- & --- & 0.108\\
\rotop (2 keyframes) \cite{li2016roto++} & .974 & .691 & --- & --- & 0.156\\
\cpc~\cite{lu2016coherent} & \textbf{.998}$^*$ & \bf{.975}$^*$ & --- & --- & --- \\
\hline
\acro: Baseline Conf.  & .993 & .932 & 0.012 & 0.152 & 0.044\\
\acro: Lean Conf.  & .995 & .938 & 0.159 & 0.541 & 0.298 \\
\acro: Medium Conf. & .995  &  .939 & 0.391  & 1.521 & 0.776\\
\acro: Full Conf.  & \underline{.995} & \underline{.951} &  1.687 & 19.21  &  8.412 \\
\hline
\end{tabular}
\label{tab:comp_iou_cpc}
\end{table}

% ---------------------------------------------------------------
\paragraph{Ablation study.} 
To evaluate the importance of each part of our model, we consider 4 different configurations:
\begin{itemize}
	\item Baseline: negative gradient with $\ell_2$-regularizer;
	\item Lean: baseline + local appearance model;
	\item Medium: lean + landmarks;
	\item Full: medium + automatic re-parametrization and global appearance model;
\end{itemize}
For all configurations, we used cross-validation (maximizing the mean IoU) on the training fold of the \davis~dataset to set the parameters and kept them fixed for all experiments.

% ---------------------------------------------------------------
\vspace{-0.4cm}
\paragraph{Quantitative results.} 
The quantitative results for the \cpc~dataset are summarized in Table \ref{tab:comp_iou_cpc}.
While average accuracy is quite similar and saturated for all methods, all configurations of \acro~outperform all baselines.  
In terms of IoU all versions of \roam, with the full configuration being the best, outperforms significantly all others.
The reason why landmarks (``medium conf.'') do not add much to \roam~is that the \cpc~dataset does not exhibit many large displacements.
The \cpc~method \cite{lu2016coherent} was evaluated only on the first ten frames of each sequence since their authors have released results only on these frames and not yet their code.
As a consequence, the average accuracy and average IoU scores that we computed and reported in Table \ref{tab:comp_iou_cpc} for \cpc~are based on partial videos, as opposed to the scores for all the other methods (including our approach).
When similarly restricted to 10 first frames, \acro~performs on par with \cpc~for all the sequences of the dataset except ``drop''. The latter sequence, which shows a water drop falling down, is of a very particular kind: the object of interest is transparent, making color models (both local and global) useless if not harmful, and exhibits a very smooth round shape. For this sequence, the \cpc~method~\cite{lu2016coherent}, which makes use of B\'ezier curves and relies solely on the strength of the image gradients, is clearly more adapted and indeed delivers better results.

\begin{figure}[tbh]
	\centering
	\begin{center}
		\setlength{\tabcolsep}{2pt}
		\begin{tabular}{ccccc}
			%\hline
			\includegraphics[width=0.10\linewidth]{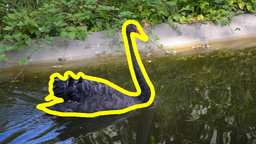} &
			\includegraphics[width=0.2\linewidth]{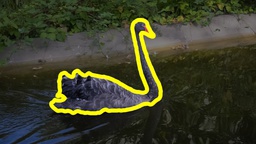} &			
			\includegraphics[width=0.2\linewidth]{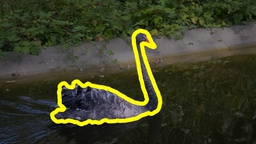} &
			\includegraphics[width=0.2\linewidth]{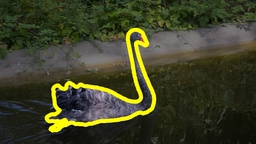} &
			\includegraphics[width=0.2\linewidth]{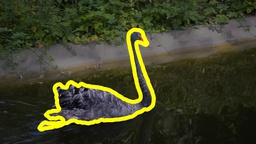} \\
			
			\includegraphics[width=0.10\linewidth]{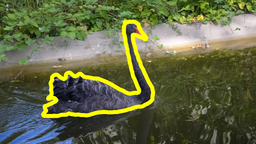} &
			\includegraphics[width=0.2\linewidth]{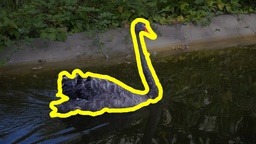} &			
			\includegraphics[width=0.2\linewidth]{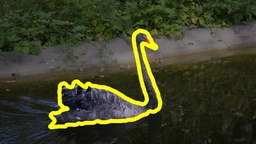} &
			\includegraphics[width=0.2\linewidth]{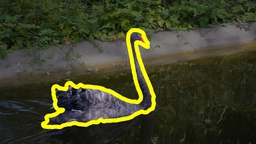} &
			\includegraphics[width=0.2\linewidth]{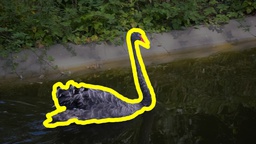} \\		
			
			\includegraphics[width=0.10\linewidth]{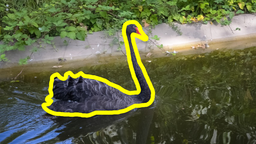} &
			\includegraphics[width=0.2\linewidth]{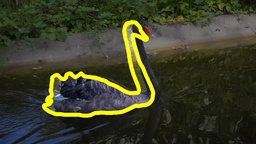} &			
			\includegraphics[width=0.2\linewidth]{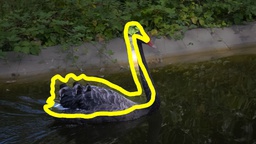} &	
			\includegraphics[width=0.2\linewidth]{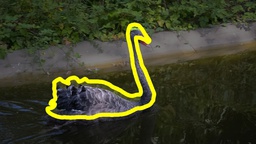} &
			\includegraphics[width=0.2\linewidth]{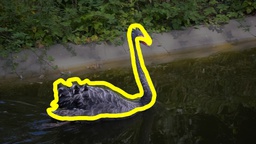} \\	
			
			\includegraphics[width=0.10\linewidth]{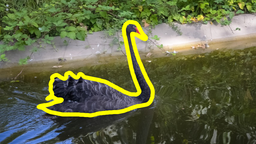} &
			\includegraphics[width=0.2\linewidth]{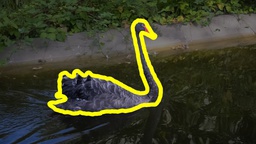} &			
			\includegraphics[width=0.2\linewidth]{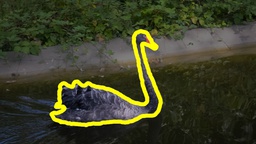} &
			\includegraphics[width=0.2\linewidth]{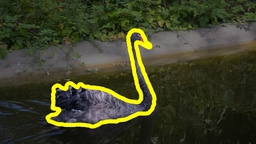} &
			\includegraphics[width=0.2\linewidth]{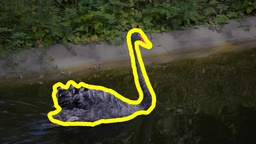} \\

			\includegraphics[width=0.10\linewidth]{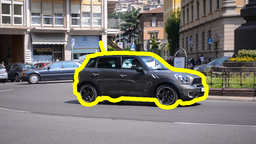} &	
			\includegraphics[width=0.2\linewidth]{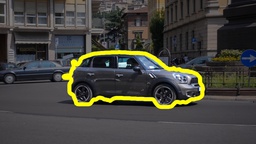} &		
			\includegraphics[width=0.2\linewidth]{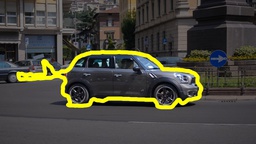} &			
			\includegraphics[width=0.2\linewidth]{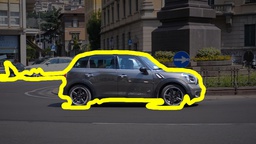} &
			\includegraphics[width=0.2\linewidth]{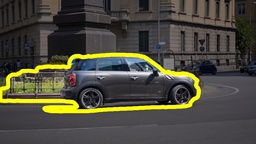} \\
			
			\includegraphics[width=0.10\linewidth]{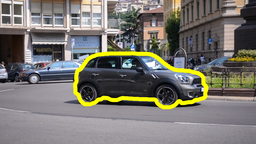} &
			\includegraphics[width=0.2\linewidth]{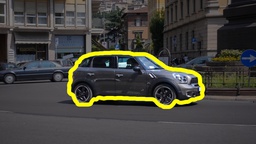} &			
			\includegraphics[width=0.2\linewidth]{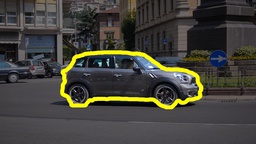} &	
			\includegraphics[width=0.2\linewidth]{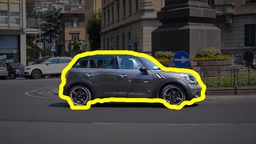} &
			\includegraphics[width=0.2\linewidth]{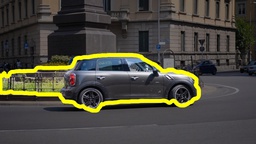} \\		
			
			\includegraphics[width=0.10\linewidth]{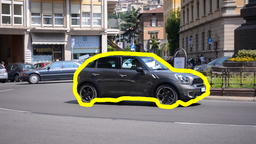} &
			\includegraphics[width=0.2\linewidth]{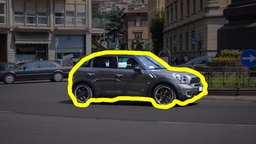} &			
			\includegraphics[width=0.2\linewidth]{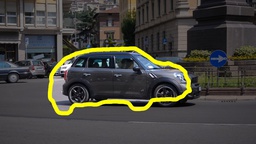} &	
			\includegraphics[width=0.2\linewidth]{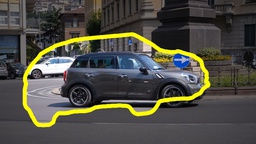} &
			\includegraphics[width=0.2\linewidth]{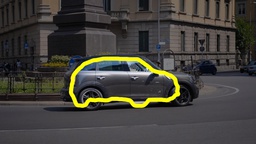} \\	
			
			\includegraphics[width=0.10\linewidth]{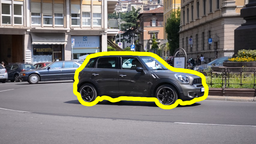} &
			\includegraphics[width=0.2\linewidth]{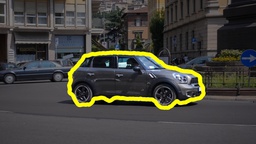} &			
			\includegraphics[width=0.2\linewidth]{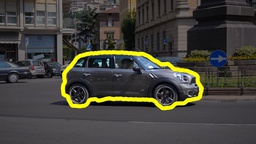} &
			\includegraphics[width=0.2\linewidth]{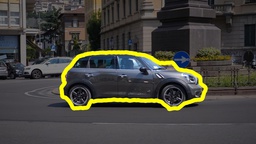} &
			\includegraphics[width=0.2\linewidth]{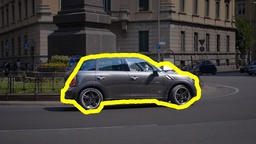} \\							
			%\hline
		\end{tabular}
	\end{center}
	\caption{{\bf Qualitative results on the \davis~dataset}: Comparisons on \textit{blackswan} and \textit{car-roundabout} sequences, between (from top to bottom for each sequence): \jump, \roto, \rotop~and \acro.
	}
	\label{fig:results_davis1}
\end{figure}

Results for the \davis~dataset are reported in Table \ref{tab:comp_iou_davis}, and qualitatively illustrated in Figs. \ref{fig:results_davis1} and \ref{fig:results_davis2}.
While our method is on par with \jump~(pixel-wise labelling), we again significantly outperform \rotop~by almost $25$ IoU points (note that using \rotop~with only two keyframes is not a typical scenario, however, this shows how complementary our approaches are).
Comparing the different configurations of \roam--local appearance models add $3$ points, landmarks $15$ and global model with re-parametrization another $5$ points--demonstrates the importance of all components of our framework.
Since it is difficult to understand the behavior of each method from average scores, we also report IoU score for each frame in Fig. \ref{fig:temp_iou}, with in addition the effect of varying the size of the displacement space in \roam~(from windows of $3\times 3$ to $13\times 13$ pixels) represented with a blue shadow. 
It can be seen that \roam~is more robust in time, not experiencing sudden performance drops as others. 
Note, that we could not provide more quantitative comparison since results/implementations of other methods were not available from the authors. 
In particular, comparison with the \cpc~method~\cite{lu2016coherent} would be interesting since the \davis~dataset \cite{perazzi2016a} exhibits many large displacements and major topology changes.

% ---------------------------------------------------------------

\begin{table}[tb]
\centering
\caption{Different types of contour warping for handling long displacements on a subset of sequences of the \davis~dataset.}
\small
\setlength{\tabcolsep}{2pt}
\begin{tabular}{c|c|c}
\hline
 & \multicolumn{1}{c|}{Average } & \multicolumn{1}{c}{Average} \\
Warping method & Accuracy &  IoU \\
\hline
Optical flow & 0.878 & 0.312 \\
Node projection from landmark tracking & 0.906 & 0.480  \\
Robust rigid transf. from landmarks  & \bf{0.934} & \bf{0.581} \\
\hline
\end{tabular}
\label{tab:comp_warp_davis}
\end{table}

\vspace{-.35cm}
\paragraph{Importance of landmarks and warping.} 
Using alternating optimization has one more benefit.
We can use the predicted position of landmarks in the next frame to estimate the transformation between the two and ``warp'' the contour to the next frame. 
This allows us to reduce the number of $D$ possible moves of nodes which i) significantly speeds-up the algorithm, ii) allows us to handle large displacement and iii) allows to better control non-rigid deformations.

We have experimented with three settings for warping of contour: using a smoothed optical flow masked as in \cite{perez2016object}, moving each node by averaging the motion of all landmarks connected to given node and robustly estimated similarity transformation with RANSAC from position of landmarks.
Table \ref{tab:comp_warp_davis} and Fig. \ref{fig:plane_landmarks} show the effect of using robustly estimate similarity transformation from position of landmarks.

\begin{figure}[tbh]
	\begin{center}
		\includegraphics[trim=0 8.8cm 4cm 0,clip,width=1\linewidth]{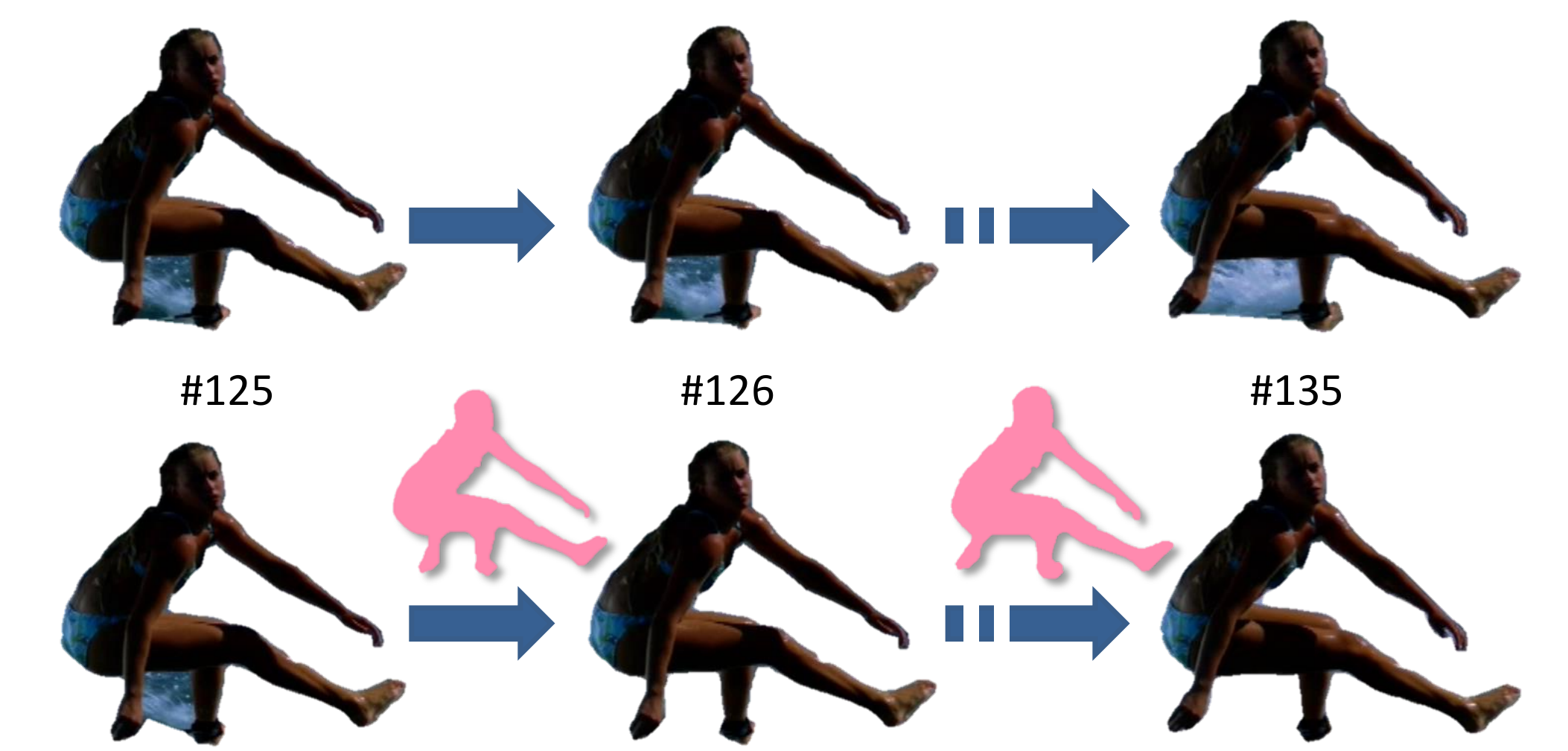}
	\end{center}
	\caption{{\bf Using proposals based on graph-cut}: Proposals (in pink) obtained through graph-cut minimization of an instrumental labeling energy using current color models allows \acro~ to monitor and accommodate drastic changes of object's outline (Bottom). Without this mechanism, parts of surrounding water get absorbed in surfer's region, between the leg and the moving arm (Top).}
	\label{fig:proposal}
\end{figure}

\begin{figure}[h]
\begin{center}
\includegraphics[trim=0 9.7cm 0 0,clip,width=1\linewidth]{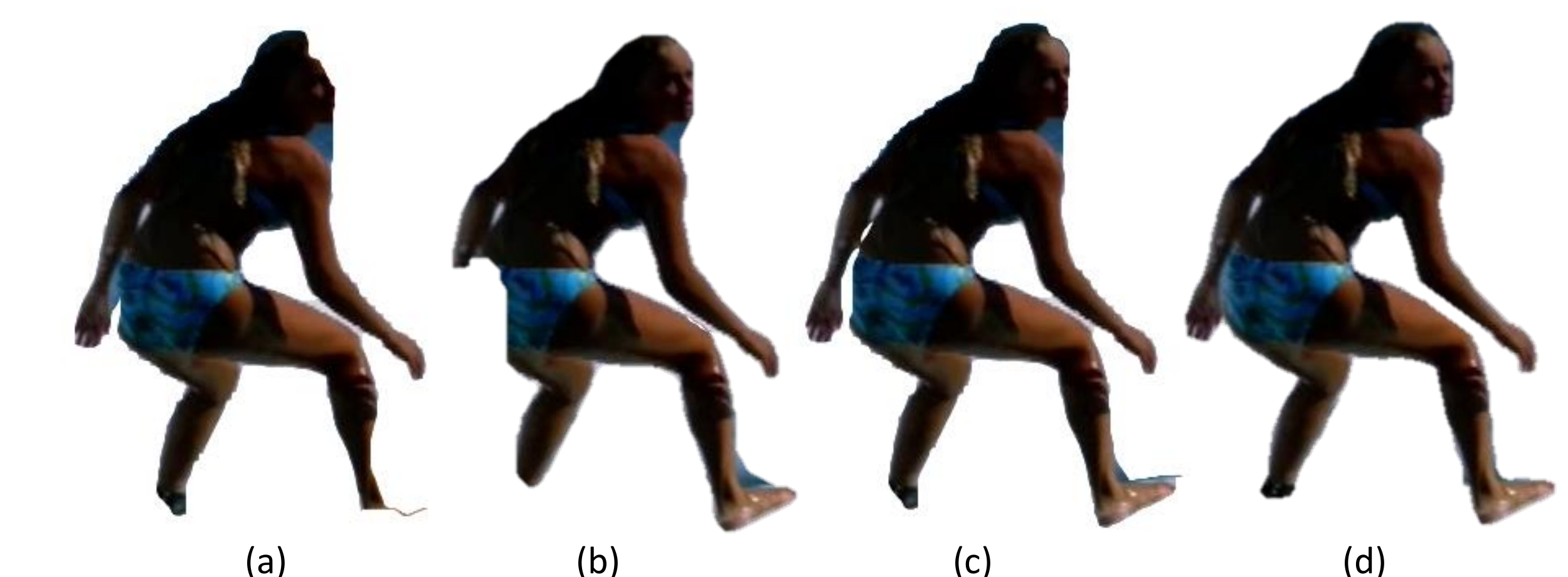}
\end{center}
%\caption{{\bf Assessing first part of the model}. (a) Edge strength only; (b) Global color model; (c) Edge strength combined with global color model; (d) With full cost function $E_C$, including local color modeling, on frame 13 from \emph{surfer} sequence.}
\caption{{\bf Assessing first part of the model}. (a) Edge strength only; (b) Global color model; (c) Edge strength combined with global color model; (d) With full cost function $E^C$, including local color modeling, on frame 13 from \emph{surfer} sequence.}
\label{fig:ingredients}
\end{figure}

\paragraph{Global color models and reparametrization.}
We investigated the effects of adding reparametrization and global color models to our framework. 
The numeric benefits of these elements can be seen in Tab.~\ref{tab:comp_iou_davis} and qualitative results on the {\it surfer} sequence from \snap~ dataset are provided in Figs.~\ref{fig:proposal} and~\ref{fig:ingredients}.
Observe that the local color models are a powerful way to capture local appearance complexities of an object through a video sequence. However, self-occlusions and articulated motion can cause these models to fail (right arm crossing the right leg of the surfer). Our contour reparametrization allows the efficient handling of this situation. Furthermore, the beneficial effect of the global color models can be observed in Fig.~\ref{fig:ingredients}, where the right foot of the surfer is successfully tracked along the whole sequence.
%the right foot of the surfer is successfully tracked along the whole sequence.

\paragraph{Timing breakdown.}
Table \ref{tab:timings_full_davis} provides detailed timing breakdown for our algorithm.
These timings were obtained on an Intel Xeon 32@3.1GHz CPU machine with 8GB RAM and Nvidia GeForce Titan X GPU.
Note that only part of the approach (evaluation of various potentials) was run on the GPU. 
In particular, both dynamic programming and re-parametrization steps could also be easily run on the graphic card, yielding real-time performance. 

\begin{table}[tbh]
	\centering
	\caption{Timing details for full configuration of \acro.}
	\small
	\setlength{\tabcolsep}{2pt}
	\begin{tabular}{c|c|c|c}
		\hline
		Step & Min. & Max. & Avg. \\
		\hline
		DP Contour   & 0.141 & 0.589 & 0.654  \\		
		DP Landmarks & 0.045 & 0.052 & 0.050  \\
		Local models edge terms & 0.326 & 0.545 & 0.522 \\ 
		Other terms & 0.012 & 0.015 & 0.013 \\
		Reparametrization & 1.093 & 8.341 & 4.101 \\
		\hline
	\end{tabular}
	\label{tab:timings_full_davis}
\end{table}

\begin{figure}[tb]
\centering
\setlength{\tabcolsep}{1pt}
\begin{tabular}{cccc}
	%\hline
\includegraphics[width=0.48\linewidth]{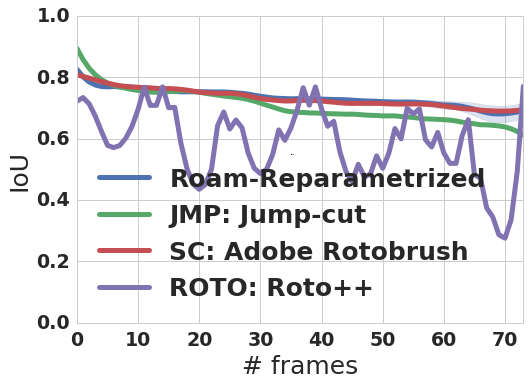} \; &
\includegraphics[width=0.48\linewidth]{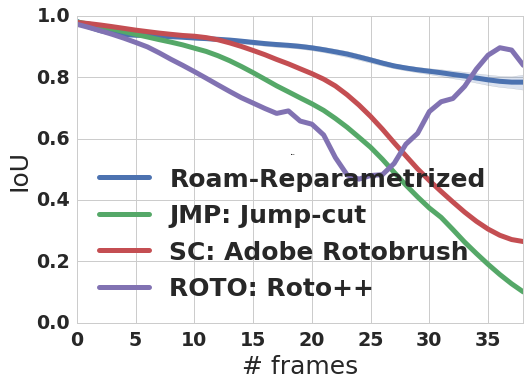}  \\
Boat & Car-shadow
	%\hline
\end{tabular}
\caption{\label{fig:temp_iou}%
	{\bf Evolution of IoU for different sequences of the \davis~dataset}. 
	For our method, the blue shadow indicates influence of varying the label space size for each node (set of possible moves in dynamic programming inference).
}
\end{figure}

\begin{figure}[tbh]
\centering
\begin{center}
\setlength{\tabcolsep}{2pt}
\begin{tabular}{ccc}
	%\hline
& %\includegraphics[width=0.15\linewidth]{fig/landmarks/plane/cont_1_no.png} & 
\includegraphics[width=0.38\linewidth]{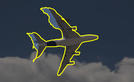}  &
\includegraphics[width=0.38\linewidth]{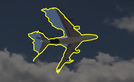}  \\
\includegraphics[width=0.2\linewidth]{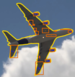} & 
\includegraphics[width=0.38\linewidth]{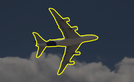} & 
\includegraphics[width=0.38\linewidth]{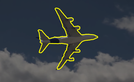}  \\
Landmarks & Frame 15 & Frame 25
	%\hline
\end{tabular}
\end{center}
\caption{{\bf Benefit of landmarks-based modeling}. Automatically detected landmarks (orange bounding boxes) are accurately tracked on the \textit{plane} sequence. This further improves the control of the boundary (bottom), as compared to \acro~without landmarks (top).
}
\label{fig:plane_landmarks}
\end{figure}

\begin{figure}[tbh]
	\centering
	\begin{center}
		\setlength{\tabcolsep}{1pt}
		\begin{tabular}{cccc}
			%\hline
			\includegraphics[width=0.33\linewidth]{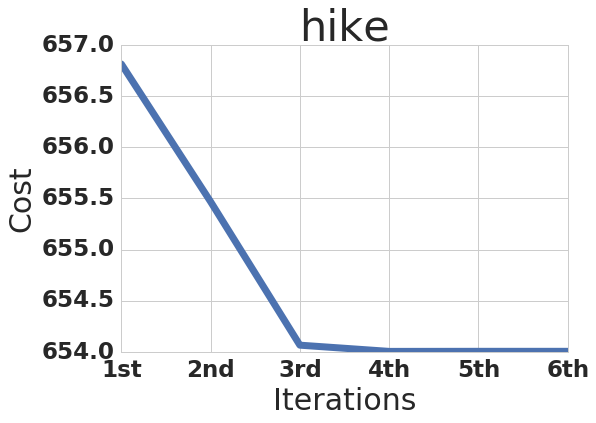} &
			\includegraphics[width=0.33\linewidth]{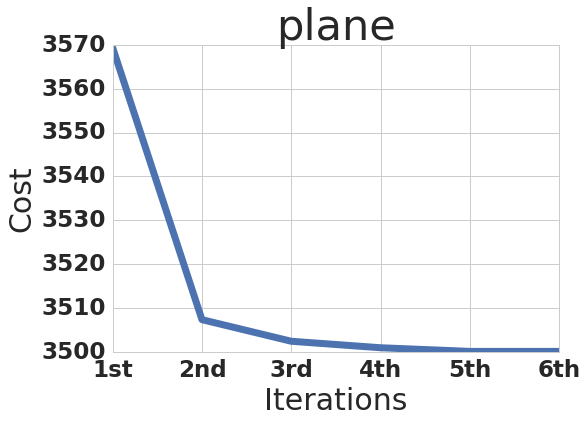} &
			\includegraphics[width=0.33\linewidth]{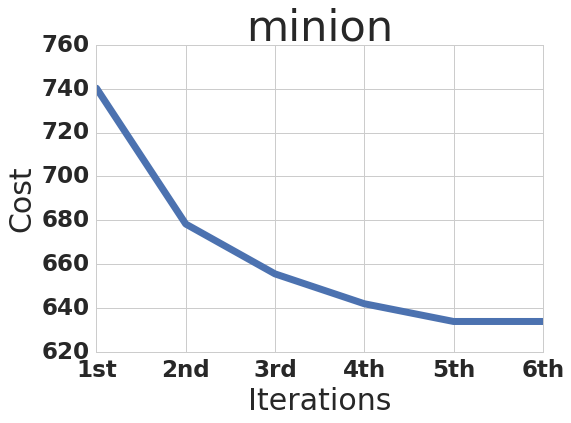} \\
			%\hline
		\end{tabular}
	\end{center}
	\caption{Energy vs. number of iterations on three sequences from the experimental datasets.}
	\label{fig:results_cost}
\end{figure}

\paragraph{Convergence.}
Fig. \ref{fig:results_cost} demonstrates that the alternating optimization described in \S \ref{sec:videocut} converges quickly within a few iterations.
	
\paragraph{Other qualitative results.} Result samples on several sequences from \davis~dataset in Figs. \ref{fig:results_davis1} and \ref{fig:results_davis2} demonstrate the superior robustness of \roam~compared to other approaches when rotoscoping of the first image only is provided as input (and last image as well for \rotop). Additional results obtained by \roam~on a variety of sequences are illustrated in Figs. \ref{fig:results_rotop} and \ref{fig:results_misc}. 

\begin{figure*}[tbh]
	\centering
	\begin{center}
		\setlength{\tabcolsep}{1pt}
		\begin{tabular}{ccccc}
			%\hline
			\includegraphics[width=0.20\linewidth]{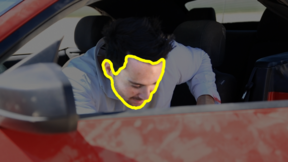} &
			\includegraphics[width=0.20\linewidth]{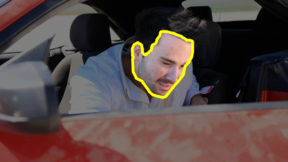} &
			\includegraphics[width=0.20\linewidth]{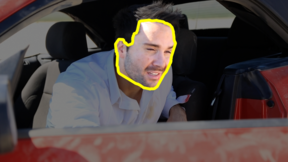} &
			\includegraphics[width=0.20\linewidth]{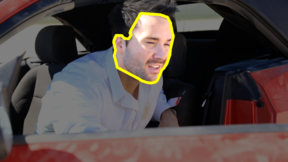} &
			\includegraphics[width=0.20\linewidth]{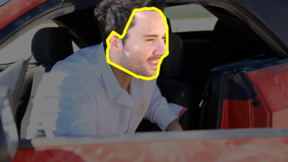} \\
			
			\includegraphics[width=0.20\linewidth]{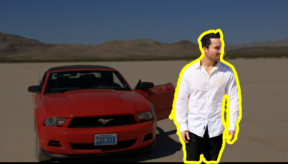} &
			\includegraphics[width=0.20\linewidth]{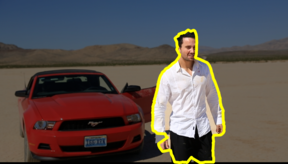} &
			\includegraphics[width=0.20\linewidth]{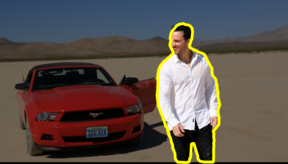} &
			\includegraphics[width=0.20\linewidth]{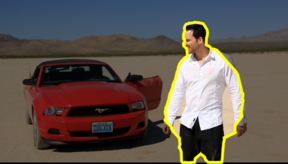} &
			\includegraphics[width=0.20\linewidth]{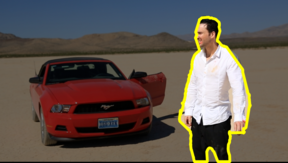} \\		
			
			\includegraphics[width=0.20\linewidth]{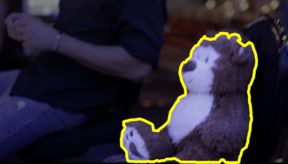} &
			\includegraphics[width=0.20\linewidth]{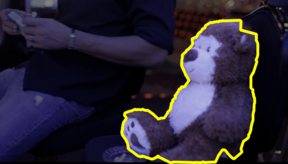} &
			\includegraphics[width=0.20\linewidth]{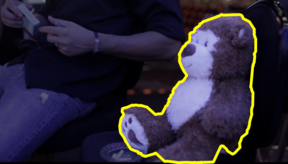} &
			\includegraphics[width=0.20\linewidth]{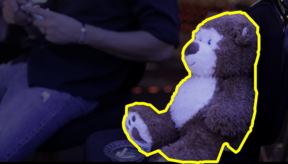} &
			\includegraphics[width=0.20\linewidth]{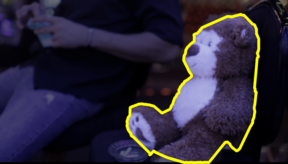} \\
			
			\includegraphics[width=0.20\linewidth]{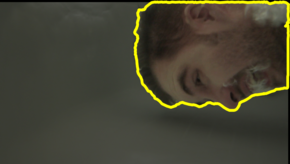} &
			\includegraphics[width=0.20\linewidth]{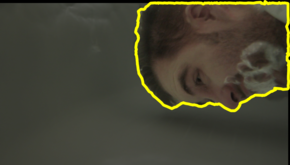} &
			\includegraphics[width=0.20\linewidth]{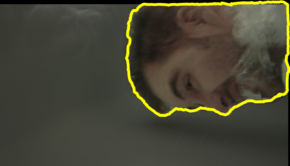} &
			\includegraphics[width=0.20\linewidth]{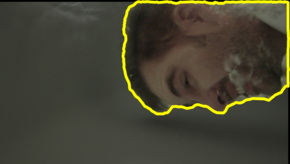} &
			\includegraphics[width=0.20\linewidth]{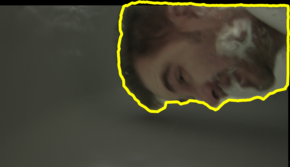} \\							
			%\hline
		\end{tabular}
	\end{center}
	\caption{Qualitative results on four sequences from the \rotop~dataset}
	\label{fig:results_rotop}
\end{figure*}

\begin{figure*}[tbh]
	\centering
	\begin{center}
		\setlength{\tabcolsep}{2pt}
		\begin{tabular}{ccccc}
			%\hline
			\includegraphics[width=0.10\linewidth]{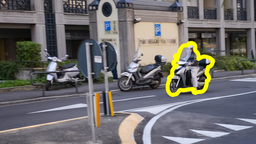} &
			\includegraphics[width=0.22\linewidth]{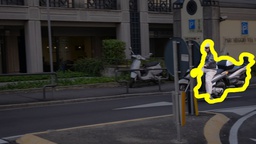} &			
			\includegraphics[width=0.22\linewidth]{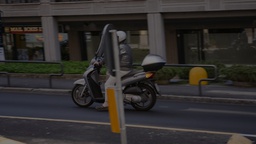} &
			\includegraphics[width=0.22\linewidth]{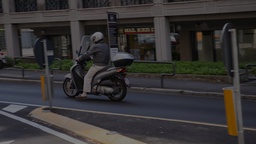} &
			\includegraphics[width=0.22\linewidth]{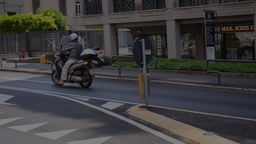} \\
			
			\includegraphics[width=0.10\linewidth]{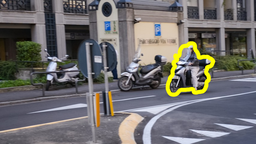} &
			\includegraphics[width=0.22\linewidth]{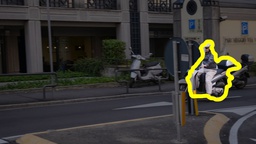} &			
			\includegraphics[width=0.22\linewidth]{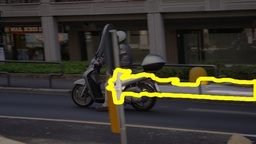} &
			\includegraphics[width=0.22\linewidth]{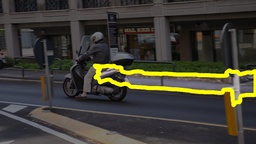} &
			\includegraphics[width=0.22\linewidth]{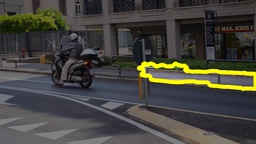} \\		
			
			\includegraphics[width=0.10\linewidth]{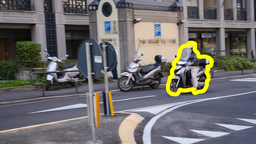} &
			\includegraphics[width=0.22\linewidth]{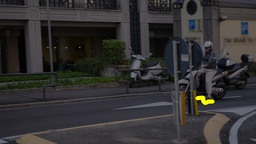} &			
			\includegraphics[width=0.22\linewidth]{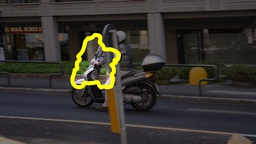} &	
			\includegraphics[width=0.22\linewidth]{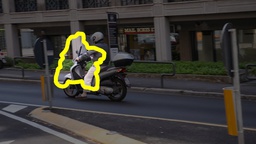} &
			\includegraphics[width=0.22\linewidth]{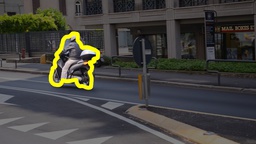} \\	
			
			\includegraphics[width=0.10\linewidth]{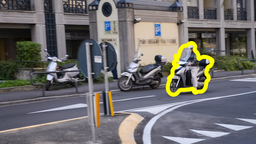} &
			\includegraphics[width=0.22\linewidth]{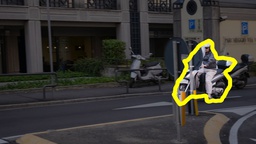} &			
			\includegraphics[width=0.22\linewidth]{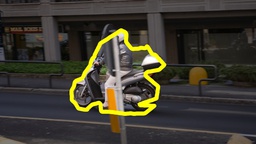} &
			\includegraphics[width=0.22\linewidth]{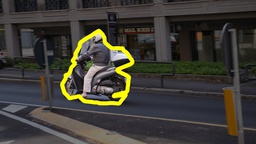} &
			\includegraphics[width=0.22\linewidth]{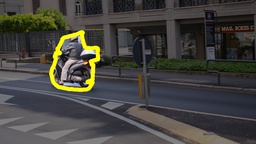} \\

			\includegraphics[width=0.10\linewidth]{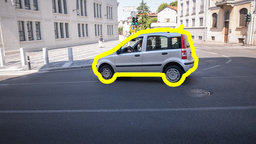} &
			\includegraphics[width=0.22\linewidth]{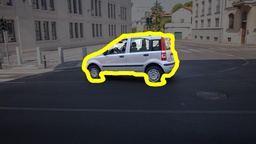} &		
			\includegraphics[width=0.22\linewidth]{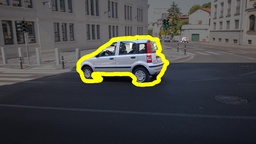} &
			\includegraphics[width=0.22\linewidth]{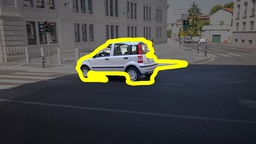} &
			\includegraphics[width=0.22\linewidth]{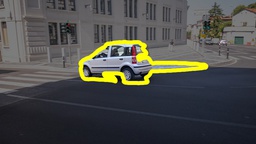} \\
			
			\includegraphics[width=0.10\linewidth]{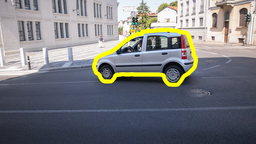} &
			\includegraphics[width=0.22\linewidth]{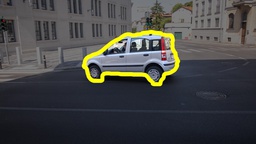} &			
			\includegraphics[width=0.22\linewidth]{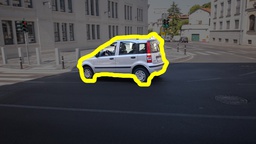} &	
			\includegraphics[width=0.22\linewidth]{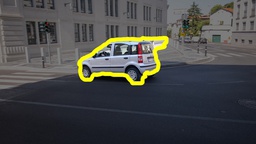} &
			\includegraphics[width=0.22\linewidth]{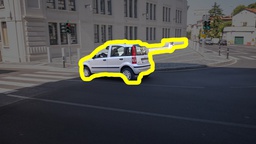} \\		
			
			\includegraphics[width=0.10\linewidth]{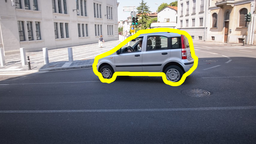} &
			\includegraphics[width=0.22\linewidth]{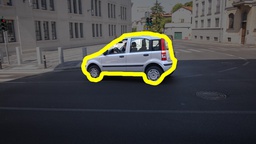} &			
			\includegraphics[width=0.22\linewidth]{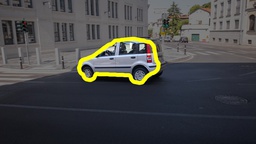} &	
			\includegraphics[width=0.22\linewidth]{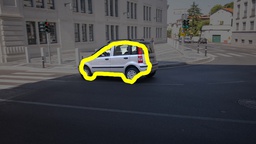} &
			\includegraphics[width=0.22\linewidth]{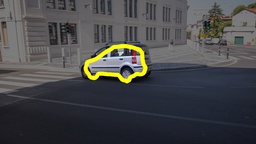} \\	
			
			\includegraphics[width=0.10\linewidth]{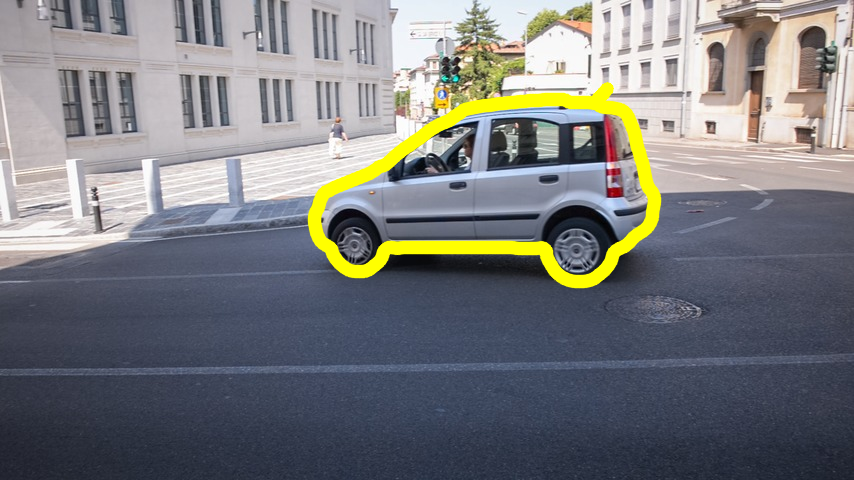} &
			\includegraphics[width=0.22\linewidth]{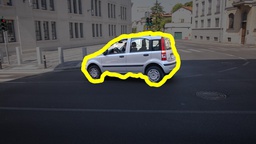} &			
			\includegraphics[width=0.22\linewidth]{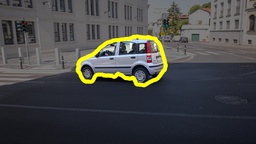} &
			\includegraphics[width=0.22\linewidth]{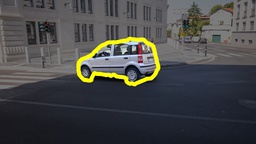} &
			\includegraphics[width=0.22\linewidth]{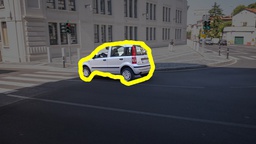} \\							
			%\hline
		\end{tabular}
	\end{center}
	\caption{{\bf More qualitative results on the \davis~dataset}: Comparisons on \textit{scooter-gray} and \textit{car-shadow} sequences between (from top to bottom): \jump, \roto, \rotop~and \acro.
	}
	\label{fig:results_davis2}
\end{figure*}

\begin{figure*}[tbh]
		\centering
		\begin{center}
			\setlength{\tabcolsep}{1pt}
			\begin{tabular}{ccccc}
				\includegraphics[width=0.19\linewidth]{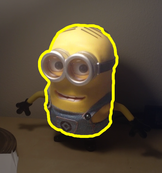} &
				\includegraphics[width=0.19\linewidth]{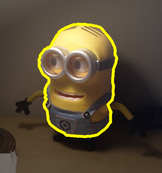} &
				\includegraphics[width=0.19\linewidth]{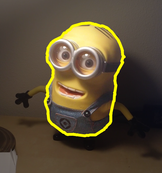} &
				\includegraphics[width=0.19\linewidth]{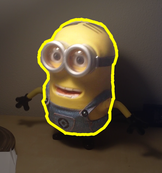} &
				\includegraphics[width=0.19\linewidth]{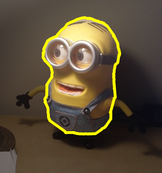}\\
				
				\includegraphics[width=0.19\linewidth]{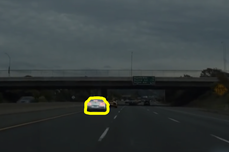} &
				\includegraphics[width=0.19\linewidth]{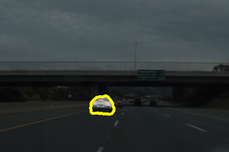} &
				\includegraphics[width=0.19\linewidth]{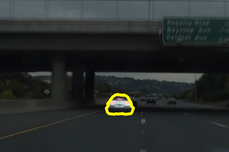} &
				\includegraphics[width=0.19\linewidth]{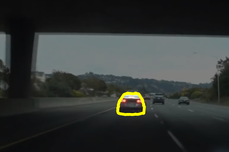} &
				\includegraphics[width=0.19\linewidth]{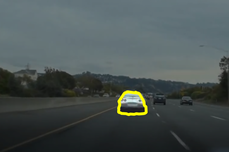}\\	
				
				\includegraphics[width=0.19\linewidth]{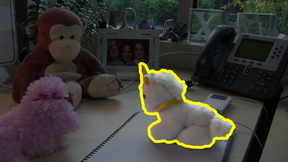} &
				\includegraphics[width=0.19\linewidth]{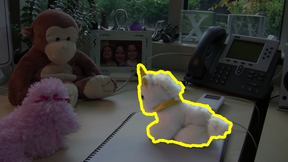} &
				\includegraphics[width=0.19\linewidth]{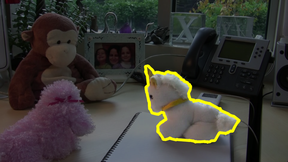} &
				\includegraphics[width=0.19\linewidth]{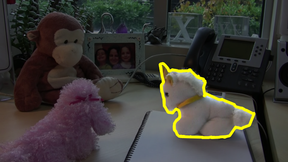} &
				\includegraphics[width=0.19\linewidth]{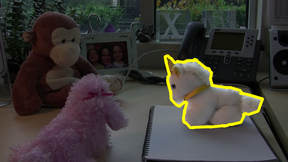}\\
				
				\includegraphics[width=0.19\linewidth]{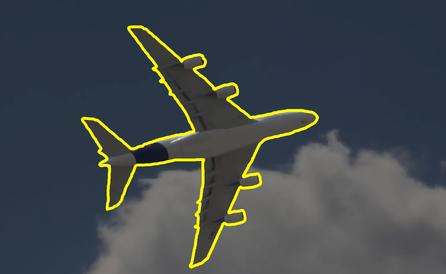} &
				\includegraphics[width=0.19\linewidth]{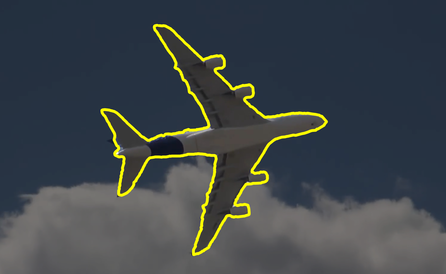} &
				\includegraphics[width=0.19\linewidth]{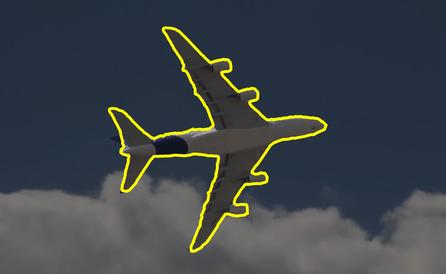} &
				\includegraphics[width=0.19\linewidth]{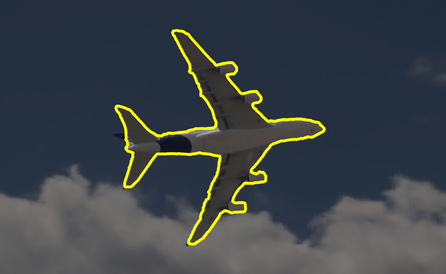} &
				\includegraphics[width=0.19\linewidth]{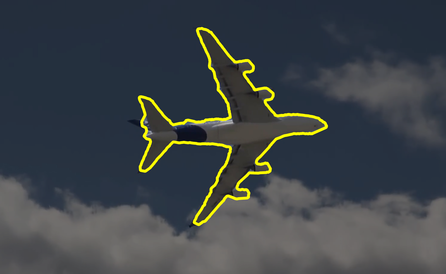}\\							
				
				\includegraphics[width=0.19\linewidth]{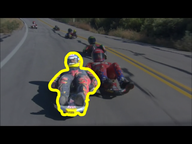} &
				\includegraphics[width=0.19\linewidth]{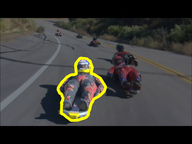} &
				\includegraphics[width=0.19\linewidth]{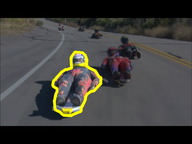} &
				\includegraphics[width=0.19\linewidth]{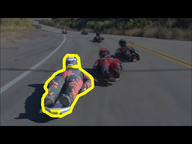} &
				\includegraphics[width=0.19\linewidth]{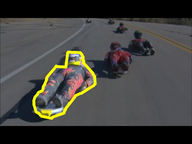}\\
				
				\includegraphics[width=0.19\linewidth]{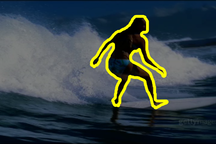} &
				\includegraphics[width=0.19\linewidth]{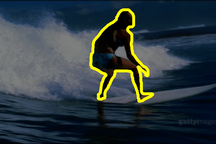} &
				\includegraphics[width=0.19\linewidth]{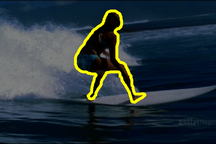} &
				\includegraphics[width=0.19\linewidth]{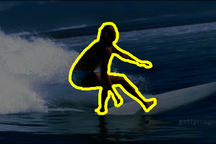} &
				\includegraphics[width=0.19\linewidth]{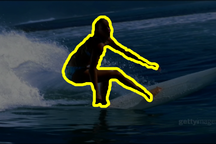}\\									
				\includegraphics[width=0.19\linewidth]{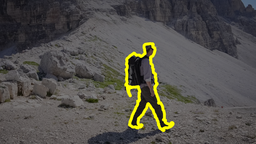} &
				\includegraphics[width=0.19\linewidth]{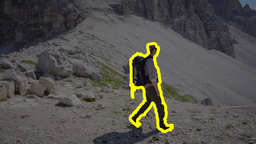} &
				\includegraphics[width=0.19\linewidth]{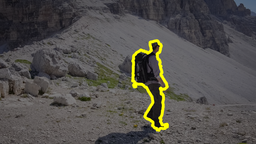} &
				\includegraphics[width=0.19\linewidth]{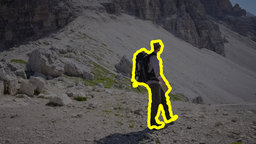} &
				\includegraphics[width=0.19\linewidth]{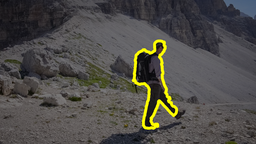}\\						
				
			\end{tabular}
		\end{center}
		\caption{\textbf{More qualitative results:} Output of \roam~on very different sequences from \davis, \cpc~and \snap~datasets among others.}
		\label{fig:results_misc}
	\end{figure*}

%% file: text/conclusion_new.tex
\section{Conclusion}
\label{sec:conclusion}

We have introduced \roam, a model to capture the appearance of the object defined by a closed curve. This model is well suited to conduct rotoscoping in video shots, a difficult task of considerable importance in modern production pipelines. We have demonstrated its merit on various competitive benchmarks.  
Beside its use within a full rotoscoping pipeline, \roam~could also be useful for various forms of object editing that require both accurate enough segmentation of arbitrary objects in videos and tracking through time of part correspondences, \eg \cite{khan2006image,rav2008unwrap}. 
Due to its flexibility, \roam~can be easily extended; in particular, with the recent \rotop~and its powerful low-dimensional shape model.